# Texture Image Classification Using DWT–AlexNet Feature Fusion and Deep Neural Networks

**Arun D. Kulkarni**
Computer Science Department,
The University of Texas at Tyler, Tyler, TX 75799 USA
e-mail: akulkarni@uttyler.edu.

**Abstract** Texture image classification plays a significant role in computer vision applications, including industrial inspection, medical image analysis, remote sensing, and object recognition. Handcrafted features can capture local texture characteristics but may have limited capability to represent complex visual patterns. In contrast, deep learning models automatically learn discriminative representations but may not fully exploit the multiscale spatial-frequency information inherent in texture images. This paper proposes a hybrid feature-fusion framework, termed DWT–AlexNet–DNN, which combines Discrete Wavelet Transform (DWT) features with deep features extracted using AlexNet for texture image classification. DWT is applied to input images to capture multiresolution spatial-frequency information from different frequency sub-bands, while AlexNet extracts high-level hierarchical visual features. These representations are fused through feature-level concatenation to form a hybrid feature vector, which is classified using a Deep Neural Network (DNN) with a SoftMax output layer. The framework is evaluated using three benchmark texture datasets: Brodatz, KTH-TIPS, and the Flickr Material Database (FMD). Each dataset is randomly partitioned into 70% training, 15% validation, and 15% testing subsets. The proposed framework achieves 100% accuracy on both Brodatz and KTH-TIPS and 88.67% on FMD. For FMD, accuracy improves from 64.67% using the AlexNet–DNN model to 88.67% using the proposed framework. The results demonstrate that feature-level fusion of DWT and AlexNet representations can substantially improve classification performance, particularly for the FMD dataset. DWT provides multiresolution spatial-frequency information, whereas AlexNet captures hierarchical and discriminative visual characteristics. Their complementary nature produces a richer feature representation and can enhance classification performance.

## 1. Introduction

Texture is an important visual property that characterizes the spatial arrangement, repetition, and distribution of intensity or color patterns within an image. Texture image classification is therefore an important problem in computer vision, with applications in industrial inspection, medical image analysis, remote sensing, material recognition, and object recognition. Unlike conventional object classification, in which shape and semantic information often provide dominant cues, texture classification relies heavily on local patterns, spatial relationships, and variations occurring at multiple scales and orientations. Consequently, effective texture classification requires feature representations capable of capturing both detailed local structures and high-level discriminative visual characteristics. Traditional texture classification methods primarily employ handcrafted features designed to describe specific statistical, structural, or frequency-domain properties of images. Common approaches include gray-level co-occurrence matrix (GLCM) features, local binary patterns (LBP), Gabor filters, and wavelet-based descriptors. These methods can effectively characterize local texture properties and provide explicit and interpretable representations. However, their effectiveness may be limited for complex textures exhibiting variations in scale, orientation, illumination, spatial arrangement, and visual appearance. Because handcrafted descriptors are based on predefined characteristics, they may also have limited capability to represent complex and high-level visual patterns.

The development of deep convolutional neural networks (CNNs) has significantly advanced image classification by enabling the automatic learning of hierarchical feature representations directly from image data. CNNs progressively learn increasingly abstract representations, ranging from low-level structures such as edges and local patterns to higher-level and more discriminative visual features. Pretrained CNN architectures, such as AlexNet, have consequently been widely used for image classification and deep feature extraction. Despite their effectiveness, CNN-based representations may not explicitly preserve detailed multiresolution spatial-frequency information that can be important for discriminating between visually similar texture patterns. The Discrete Wavelet Transform (DWT) provides a complementary representation by decomposing an image into multiple frequency sub-bands at different spatial resolutions. Through this decomposition, DWT captures spatial-frequency information and separates image content into low- and high-frequency components. Such multiresolution analysis is particularly useful for characterizing texture patterns that exhibit variations across different scales and orientations. However, DWT-based features alone may not adequately capture the high-level hierarchical visual representations learned by deep neural networks. Thus, DWT and CNN-based representations offer complementary information that can potentially be exploited through feature fusion.

Motivated by these complementary characteristics, this paper proposes a hybrid feature-fusion framework, termed DWT–AlexNet–DNN, for texture image classification. The proposed framework consists of two parallel feature-extraction paths. In the first path, the Discrete Wavelet Transform is applied to the input images to extract multiresolution spatial-frequency features from different frequency sub-bands. In the second path, AlexNet is employed to extract high-level hierarchical visual features from the same input images. The DWT and AlexNet feature vectors are subsequently combined through feature-level concatenation to construct a unified hybrid representation. The resulting fused feature vector is classified using a Deep Neural Network (DNN) with a SoftMax output layer. The proposed framework is evaluated on three benchmark texture datasets: Brodatz, KTH-TIPS, and the Flickr Material Database (FMD). For each dataset, the images are randomly divided into 70% training, 15% validation, and 15% testing subsets. To evaluate the contribution of feature fusion, the proposed DWT–AlexNet–DNN framework is compared with classification approaches based on individual feature representations. The experimental results demonstrate that the proposed fusion strategy improves classification performance. The main contribution of this work is the development and experimental evaluation of a hybrid texture representation that combines complementary handcrafted and deep feature representations. DWT contributes multiresolution spatial-frequency information from different wavelet sub-bands, while AlexNet provides hierarchical and discriminative visual features learned from image data. Their feature-level integration produces a richer representation of texture images and improves classification performance across datasets with different visual characteristics.

The remainder of this paper is organized as follows. Section 2 presents a review of related work on handcrafted texture descriptors, deep learning-based methods, feature-fusion approaches, and benchmark texture datasets. Section 3 describes the proposed DWT–AlexNet–DNN framework, including the feature extraction, feature fusion, and classification procedures. Section 4 presents the experimental set-up and results, followed by a discussion of the findings. Section 5 concludes the paper and discusses potential directions for future research.

## 2. Related Work

Texture classification is an important research problem in computer vision and pattern recognition because texture provides valuable information about material properties, surface structure, and the spatial organization of objects and scenes. Traditional texture-classification methods characterize images using handcrafted descriptors based on statistical, structural, local, or frequency-domain properties. More recently, convolutional neural networks (CNNs) have enabled data-driven representation learning by automatically extracting hierarchical features directly from images. However, texture classification differs from conventional object-recognition tasks because texture categories are often characterized by repetitive local patterns, spatial-frequency distributions, orientation, scale, and fine-grained appearance rather than by global object shape. Consequently, both handcrafted and deep representations continue to be investigated for texture classification. Liu et al. presented a comprehensive review of the evolution of texture representations from Bag-of-Visual-Words approaches to CNN-based methods and discussed commonly used benchmark datasets and evaluation protocols [1].

Early texture-classification methods relied primarily on handcrafted statistical and structural descriptors. Haralick et al. introduced gray-level co-occurrence matrix (GLCM)-based texture features, including contrast, correlation, energy, homogeneity, and entropy, for image classification [2]. These features characterize the spatial relationships among gray-level values and have become widely used for statistical texture analysis. Their principal advantages are interpretability and low computational complexity. However, GLCM features can be sensitive to image rotation, scale, quantization, and the selection of displacement and orientation parameters. Filter-bank approaches subsequently provided multiscale and multidirectional representations for texture analysis. Gabor filters are particularly effective because they provide joint localization in the spatial and frequency domains. Jain and Farrokhnia investigated texture segmentation using Gabor filters at multiple scales and orientations [3]. Such representations can effectively capture periodic structures and directional patterns; however, the resulting feature vectors may become large and computationally expensive when numerous scales and orientations are employed. Local Binary Pattern (LBP), proposed by Ojala et al., became another influential handcrafted descriptor for texture classification [4]. LBP characterizes local spatial structure by comparing neighboring pixels with a central pixel and encoding the resulting relationships as binary patterns. Its simplicity, computational efficiency, and relative robustness to monotonic illumination changes have made it attractive for texture analysis. Several extensions of LBP have subsequently introduced rotation invariance, multiscale processing, and improved local-contrast representations. Liu et al. provided a taxonomy and experimental study of local binary features and demonstrated their competitiveness across a variety of texture datasets and imaging conditions [5]. Frequency-domain and multiresolution approaches constitute another important category of handcrafted texture descriptors. The Discrete Wavelet Transform (DWT) provides simultaneous spatial and frequency localization by decomposing an image into approximation and detail sub-bands at multiple scales. Mallat established the theoretical framework for multiresolution wavelet decomposition [6]. In texture analysis, wavelet coefficients can characterize coarse structures as well as horizontal, vertical, and diagonal details. Arivazhagan and Ganesan investigated texture classification using wavelet statistical and wavelet co-occurrence features and demonstrated the effectiveness of combining information from multiple wavelet sub-bands [7]. Wavelet-based representations are particularly suitable for texture analysis because texture information frequently occurs at multiple spatial scales. Fine texture details are represented in high-frequency sub-bands, whereas larger-scale structural information is represented by lower-frequency components. Statistical measures such as mean, variance, energy, entropy, and standard deviation can therefore be computed from different wavelet sub-bands to construct compact feature representations. Despite their advantages, handcrafted methods require careful selection and tuning of descriptors, scales, orientations, and statistical measures. Their generalization capability can consequently be limited when texture patterns become highly diverse. Overall, handcrafted descriptors provide explicit and interpretable representations of local and frequency-domain texture characteristics, but their representations are designed manually rather than learned from data. These limitations motivated the development of deep learning approaches capable of automatically learning hierarchical and discriminative texture representations.

The introduction of CNNs significantly changed image classification by enabling feature extraction and classification to be learned jointly from training data. CNNs learn hierarchical representations in which early layers capture local structures and edges, whereas deeper layers encode increasingly complex patterns. Although CNNs were initially developed primarily for object recognition, their learned representations have demonstrated considerable potential for texture and material recognition. Cimpoi et al. investigated CNN-based representations for texture and material recognition by treating convolutional layers as learned filter banks and combining the resulting features with Fisher-vector encoding [8]. Their results demonstrated that convolutional features could provide highly effective representations for texture recognition, particularly when local features are aggregated using orderless representations. Song et al. proposed discriminative neural-network representations specifically for texture image classification and demonstrated the advantages of learned CNN features over conventional handcrafted descriptors [9]. Their work also indicated that conventional CNN representations developed for object recognition may not fully capture the statistical and orderless characteristics of texture. The Deep Texture Encoding Network (Deep-TEN), introduced by Zhang et al., incorporated an encoding layer into a CNN architecture to jointly learn a visual dictionary and an orderless encoding of local features [10]. Unlike conventional CNN classification architectures, Deep-TEN explicitly models the distribution of local patterns, making it particularly suitable for texture and material recognition. The availability of pretrained CNN models has further increased the practical application of deep features to texture classification. Instead of training an entire CNN from scratch, features extracted from networks pretrained on large image datasets can be used as generic descriptors and subsequently classified using support vector machines or other machine-learning classifiers. Such transfer-learning approaches are particularly useful when texture datasets are small. Studies using pretrained CNN architectures have reported competitive performance for texture classification [11]. A major advantage of CNNs is their ability to learn discriminative features automatically, thereby reducing the need for manually designed descriptors. Nevertheless, conventional CNN architectures primarily operate in the spatial domain and do not explicitly impose a multiscale frequency-domain representation on the input. This limitation is particularly relevant to texture classification because frequency, scale, and orientation are important characteristics of texture. CNNs may also require substantial training data and computational resources, and their learned representations may not explicitly preserve fine-grained frequency information. These observations suggest that handcrafted frequency-domain descriptors and CNN-based representations can provide complementary information. Wavelet-based features explicitly characterize multiscale frequency properties, whereas CNN features provide learned hierarchical representations. This complementarity motivates the investigation of hybrid feature-fusion approaches for texture classification.

Feature fusion seeks to combine complementary representations so that information not adequately captured by one feature-extraction technique can be complemented by another. Fusion can be performed at the input, feature, decision, or classifier level. Among these approaches, feature-level fusion is particularly attractive because heterogeneous descriptors can be concatenated into a unified representation before classification. Cimpoi et al. demonstrated that deep convolutional features can be combined with orderless encoding techniques such as Fisher vectors to obtain effective texture representations [8]. Similarly, Deep-TEN integrates dictionary learning, feature encoding, and CNN representation learning into a unified architecture [10]. These approaches demonstrate the importance of modeling the distribution of local patterns for texture recognition. Another research direction incorporates frequency-domain processing directly into CNN architectures. Fujieda et al. proposed Wavelet Convolutional Neural Networks (Wavelet CNNs), in which wavelet transforms are incorporated into convolutional processing [12]. Their approach was motivated by the observation that conventional CNN operations do not explicitly exploit spectral information that can be useful for texture representation. Wavelet-based CNN architectures therefore attempt to combine the multiscale and frequency-selective properties of wavelets with the representation-learning capability of CNNs. Recent studies have continued to investigate the integration of wavelet transforms and deep learning. Gowthaman and Das proposed a dual-tree complex wavelet convolutional neural network framework that exploits directional and phase information for texture classification [13]. Park et al. introduced a deep feature retention module network that preserves information from multiple levels of a deep network for texture classification [14]. Yu et al. proposed a texture classification network integrating an adaptive wavelet transform with deep learning, demonstrating the potential of incorporating wavelet-based information into learned representations [15]. A recent systematic review by Wu et al. further examined the integration of wavelet transforms and deep neural networks and discussed their complementary characteristics and synergistic architectures [16]. Other studies have investigated the use and aggregation of deep features for texture classification. Mandal et al. evaluated ImageNet-pretrained CNNs for texture-based rock classification and demonstrated the effectiveness of general-purpose CNN representations for texture-related classification tasks [17]. Dong et al. investigated attention-based deep networks for texture classification [18], whereas Neshov et al. explored the aggregation of features from multiple layers of pretrained vision architectures [19]. Gupta et al. proposed HyTexNet, which combines local encoding and deep feature information for enhanced texture classification [20]. An alternative to modifying the internal architecture of a CNN is to extract handcrafted and deep features independently and combine them at the feature level. In this approach, wavelet features are extracted directly from the input image, while CNN features are obtained from a pretrained or trained CNN. The resulting feature vectors can then be normalized and concatenated before being supplied to a classifier. This strategy has several advantages. First, the wavelet branch provides an explicit and interpretable representation of multiscale frequency information. Second, the CNN branch provides learned nonlinear representations of texture structures. Third, feature-level fusion allows both representations to be exploited simultaneously without requiring substantial modifications to the CNN architecture. The existing literature therefore indicates that wavelet transforms and CNNs provide complementary representations for texture classification. Wavelet transforms provide localized frequency information at multiple scales, whereas CNNs automatically learn discriminative spatial representations. However, the literature provides comparatively less emphasis on a straightforward framework in which independently extracted multilevel DWT features and CNN features are directly fused and subsequently classified using a deep neural network. This observation motivates the hybrid framework proposed in this paper, in which DWT-based features and CNN-derived features are concatenated to form a unified representation for texture classification.

Reliable evaluation of texture-classification methods requires benchmark datasets containing variations in texture type, scale, illumination, viewpoint, and material appearance. Several datasets have been widely used to evaluate texture representations. The Brodatz dataset is one of the classical texture databases and contains a wide variety of grayscale texture patterns. It has historically been used to evaluate handcrafted descriptors and texture-analysis algorithms. The Columbia-Utrecht Reflectance and Texture (CUReT) dataset contains 61 material classes and approximately 5,600 images and was designed to provide variations in viewing and illumination conditions [21]. CUReT is therefore useful for evaluating the robustness of texture descriptors under controlled but varying imaging conditions. The UIUC Texture Database contains 25 texture categories and 1,000 images, with substantial variations in scale and viewpoint [22]. It has been widely used for evaluating texture representations under nonuniform imaging conditions. The KTH-TIPS dataset contains 10 material categories and 810 images acquired under variations in scale and imaging conditions. Its extension, KTH-TIPS-2b, contains 11 material categories and provides more images per class, making it an important benchmark for texture and material recognition. The Describable Textures Dataset (DTD), introduced by Cimpoi et al., contains 47 texture categories described using human-interpretable attributes such as cracked, dotted, fibrous, marbled, and wrinkled [23]. Unlike traditional material datasets, DTD emphasizes perceptual texture attributes and provides a challenging benchmark for general texture recognition. The Flickr Material Database (FMD) contains images from 10 material categories and represents materials under realistic imaging conditions. It has been widely used for evaluating material and texture recognition methods. Other commonly used databases include the UMD Texture Database, which contains 25 classes and approximately 1,000 images, and the ALOT database, which provides a larger collection with 250 categories and approximately 25,000 images [1]. More recently, T1K+ was developed as a large-scale benchmark for color texture classification and retrieval. It contains more than 1,000 texture categories and provides greater class diversity than many traditional texture databases [24].

The literature demonstrates a clear progression from handcrafted texture descriptors to learned deep representations. Statistical descriptors such as GLCM, local descriptors such as LBP, and frequency-domain approaches such as DWT provide compact and interpretable representations, but their effectiveness depends strongly on manually selected features and parameters. CNNs, in contrast, automatically learn hierarchical representations and have achieved robust performance on challenging texture and material datasets. However, conventional CNNs do not explicitly impose a multiscale frequency-domain representation on the input. Wavelet-based CNN architectures address this limitation by incorporating wavelet processing into the network itself [12], [13]. Such approaches require specialized network architectures and modifications to conventional CNN processing. An alternative strategy is to preserve the advantages of both representations by independently extracting wavelet and CNN features and subsequently combining them through feature-level fusion. Accordingly, the framework proposed in this paper employs two complementary feature-extraction paths. The first path extracts multilevel DWT-based features that explicitly characterize the spatial-frequency properties of texture images. The second path extracts deep features from a pretrained AlexNet model. The resulting feature vectors are subsequently concatenated to form a fused representation. In the proposed implementation, the DWT branch provides a 192-dimensional feature vector, while AlexNet provides a 256-dimensional feature vector, resulting in a 448-dimensional fused representation. The fused feature vector is then supplied to a deep neural network classifier with a SoftMax output layer. The underlying hypothesis is that the complementary information contained in the wavelet and CNN representations can provide a more discriminative feature space than either representation alone. This approach is particularly appropriate for texture classification because texture patterns are simultaneously characterized by local spatial structures and frequency distributions. Thus, combining handcrafted multiresolution information with learned deep representations provides a theoretically motivated and straightforward alternative to specialized wavelet-CNN architectures. The proposed framework and its implementation are described in the following section.

## 3. Proposed Approach

To investigate the effectiveness of feature fusion using two complementary techniques—Discrete Wavelet Transform (DWT) and AlexNet features—three classification models were implemented. The first model extracts DWT-based features and classifies them using a DNN and is referred to as the DWT-DNN model. The second model uses the modified AlexNet to extract a compact deep feature representation, which is subsequently classified using a DNN; this model is referred to as the AlexNet-DNN model. The third model performs feature-level fusion by combining the DWT-based features with the deep features extracted from the modified AlexNet. The resulting fused feature representation is classified using a deep neural network (DNN), and this model is referred to as the DWT-AlexNet-DNN model. These three models enable a direct comparison of the classification performance obtained using the fused feature representation with that obtained using the individual feature representations. All three models use the same dataset partitioning and evaluation protocol to ensure a consistent and fair comparison. The input to each model consists of texture images resized to 256 × 256 pixels. Let the input image be denoted by $f(x, y)$, where $x$ and $y$ represent the spatial coordinates. For each input image, two complementary feature representations are generated: a multiresolution frequency-domain representation obtained using the DWT and a learned deep feature representation extracted using the modified AlexNet. The DWT captures texture characteristics across multiple spatial-frequency scales, whereas the modified AlexNet learns discriminative features directly from the image data. In the proposed hybrid framework, these two feature representations are concatenated at the feature level to form a unified feature vector, which is subsequently classified using a DNN.

### 3.1 DWT-DNN Model

The DWT-DNN model extracts statistical texture features from a three-level two-dimensional DWT decomposition. At each decomposition level, the image is decomposed into four sub-bands: approximation (LL), horizontal detail (LH), vertical detail (HL), and diagonal detail (HH). This multilevel decomposition provides localized spatial-frequency information at different resolutions, making DWT particularly effective for characterizing texture patterns and their structural variations.

$$I(x,y) \rightarrow \{LL_l, LH_l, HL_l, HH_l\}, \quad l = 1,2,3 \qquad (1)$$

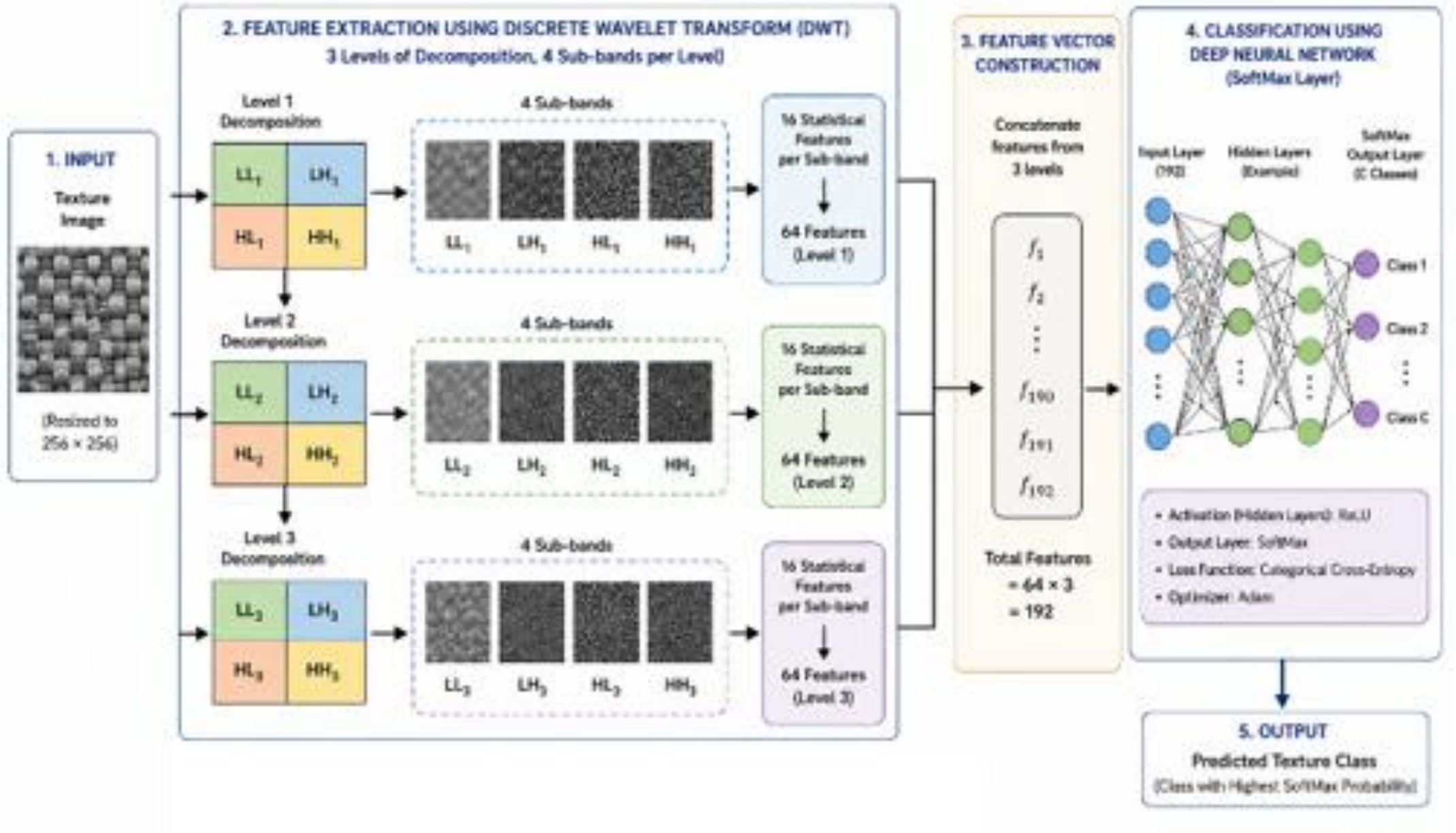

**Fig. 1.** Framework of the DWT-DNN model

Sixteen statistical descriptors are computed from each sub-band. Therefore, each decomposition level produces 4 × 16 = 64 features. With three decomposition levels, the total number of DWT features is 3 × 4 × 16 = 192. The DWT feature vector for an image is consequently represented as

$$\mathbf{f}_{DWT} = \left[\mathbf{f}_1, \mathbf{f}_2, \mathbf{f}_3\right]^T \in \mathbb{R}^{192} \qquad (2)$$

where $\mathbf{f}_l$ denotes the 64-dimensional feature vector obtained from the four sub-bands at level *l*. The 192-dimensional vector is supplied to a DNN consisting of fully connected layers followed by a SoftMax output layer. The DWT-DNN model establishes a baseline for evaluating the contribution of multiresolution handcrafted texture information. The DWT-DNN model is shown in Fig. 1.

### 3.2 ALEXNET-DNN Model

The AlexNet-DNN model employs a modified AlexNet architecture to learn discriminative representations directly from the input texture images. The convolutional layers progressively transform the input image into higher-level feature maps, while the fully connected layers provide a compact representation suitable for classification. In the proposed implementation, the network is configured to produce a 256-dimensional feature vector before the final SoftMax classification layer.

$$\mathbf{f}_{Alex} = g_{Alex\left(I;\theta_{Alex}\right)} \in \mathbb{R}^{256} \quad (3)$$

where $g_{\mathrm{Alex}\left(I;\theta_{\mathrm{Alex}}\right)}$ denotes the pretrained AlexNet feature-extraction mapping and $\theta_{Alex}$ represents its learned parameters. The 256-dimensional representation is used as the deep feature vector for each image. A SoftMax layer is used during network training to assign each image to one of the ten texture classes. The AlexNet-DNN model provides a second baseline and allows the discriminative capability of learned deep features to be assessed independently of the DWT features. The AlexNet-DNN model is shown in Fig. 2.

### 3.3 DWT-AlexNet-DNN Model

The proposed hybrid framework combines the complementary information contained in the DWT and AlexNet representations. The DWT branch captures multiresolution frequency characteristics and statistical texture variations, whereas the AlexNet branch learns higher-level spatial and visual patterns. Because the two feature vectors describe the same image using different representations, they can be concatenated at the feature level.

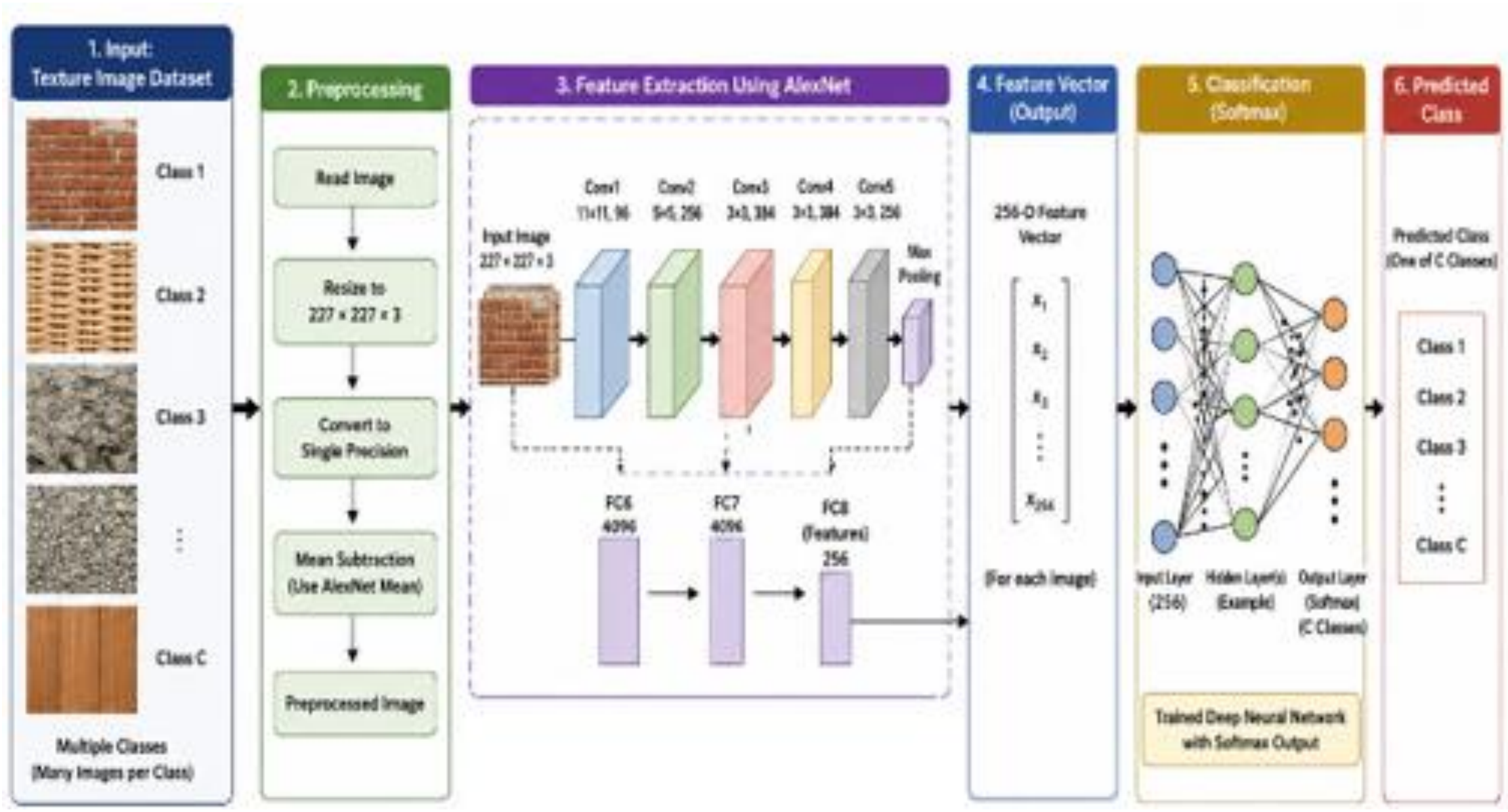

**Fig. 2.** Framework of the AlexNet-DNN model

$$\mathbf{f}_F = \left[ \mathbf{f}_{DWT}^T \, \mathbf{f}_{Alex}^T \right]^T \qquad (4)$$

**Table 1.** Layers of AlexNet

| Layer | Type | Output / Function |
|---|---|---|
| Input | Image Input | 227 × 227 × 3 RGB image |
| Conv1 | Convolution | 96 × 55 × 55 |
| ReLU1 | ReLU | Nonlinear activation |
| Pool1 | Max Pooling | 96 × 27 × 27 |
| Conv2 | Convolution | 256 × 27 × 27 |
| ReLU2 | ReLU | Nonlinear activation |
| Pool2 | Max Pooling | 256 × 13 × 13 |
| Conv3 | Convolution | 384 × 13 × 13 |
| ReLU3 | ReLU | Nonlinear activation |
| Conv4 | Convolution | 384 × 13 × 13 |
| ReLU4 | ReLU | Nonlinear activation |
| Conv5 | Convolution | 256 × 13 × 13 |
| ReLU5 | ReLU | Nonlinear activation |
| Pool5 | Max Pooling | 256 × 6 × 6 |
| FC1 | Fully Connected | 4096 |
| ReLU6 | ReLU | — |
| Dropout | Dropout | Regularization |
| FC2 | Fully Connected | 256 |
| ReLU7 | ReLU | — |
| Feature Vector | — | 256-dimensional |
| FC3 | Fully Connected | Number of classes |
| SoftMax | SoftMax | Class probabilities |
| Classification | Classification Output | Predicted class |

Since $\mathbf{f}_{\mathrm{DWT}} \in \mathbb{R}^{192}$ and $\mathbf{f}_{\mathrm{Alex}} \in \mathbb{R}^{256}$ the fused representation has 192 + 256 = 448 dimensions:

$$\mathbf{f}_F \in \mathbb{R}^{448} \qquad \textbf{(5)}$$

The fused vector is then provided to a DNN classifier composed of fully connected layers followed by a SoftMax layer. Unlike a decision-level fusion scheme, the proposed approach allows the classifier to learn interactions among the DWT and deep features jointly. Consequently, the classifier can exploit information that may not be sufficiently represented by either feature set alone. For the DWT-DNN, AlexNet-DNN, and DWT-AlexNet-DNN models, classification is performed using fully connected layers followed by a SoftMax layer. Let $z_k$ denote the output of the final fully connected layer for class *k*. The SoftMax probability assigned to class *k* is given by

$$P(y=k \mid \mathbf{f}) = \frac{e^{z_k}}{\sum_{j=1}^{C} e^{z_j}}, \; k=1,\ldots,C \qquad (6)$$

where *C* is the number of texture classes and $\mathbf{f}$ denotes the input feature vector to the classifier. The predicted class is obtained from the class having the largest posterior probability:

$$\hat{y} = \arg\ \max_k \left(P(y=k \mid \mathbf{f})\right) \qquad (7)$$

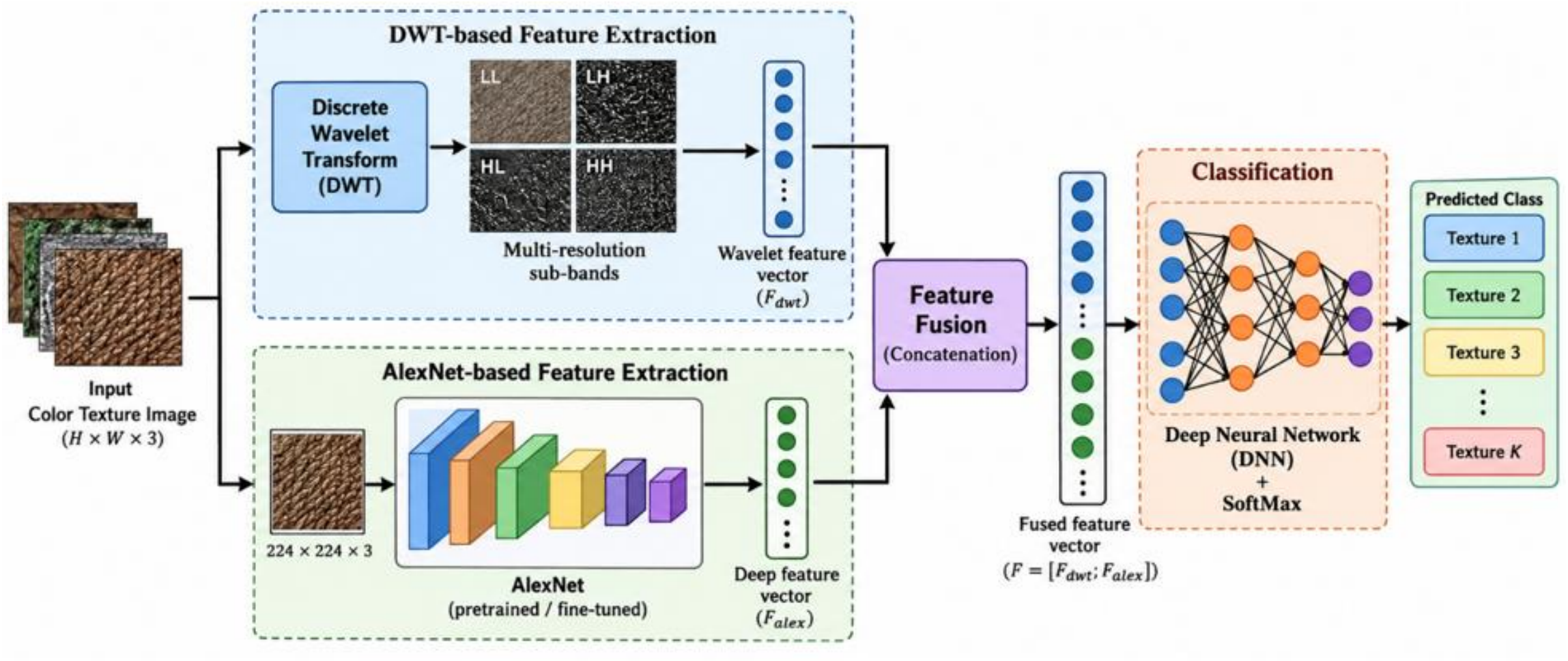

**Fig. 3.** Framework of the DWT-AlexNet-DNN model

During training, the networks are optimized using categorical cross-entropy. For a sample with one-hot encoded target vector y, the loss is defined as

$$L_{CE} = -\sum_{k=1}^{C} y_k \log\left(\hat{y}_k\right) \qquad (8)$$

For $N$ samples the average training loss is given by

$$L_{CE} = -\frac{1}{N}\sum_{i=1}^{N}\sum_{k=1}^{C} y_{ik} \log\left(\hat{y}_{ik}\right) \qquad (9)$$

Minimizing this loss encourages the network to assign high probability to the correct texture class while reducing the probability assigned to incorrect classes. The DWT AlexNet-DNN model is shown in Fig. 3. The layers of the DNN are shown in Table 2. The number of neurons in the final fully connected and SoftMax layers were set according to the number of classes in each dataset.

### 3.4 Training, Validation, and Testing Protocol

To ensure a fair comparison among the three models, the dataset is partitioned using the same 70%–15%–15% protocol for training, validation, and testing, respectively. The training set is used to optimize the network parameters, while the validation set is used to monitor training and support model selection. After training, the independent test set is used only for final performance evaluation.

$$D = D_{train} \cup D_{val} \cup D_{test}, \quad \text{with } |D_{train}|:|D_{val}|:|D_{test}| = 70:15:15 \tag{10}$$

For each test image, the predicted class is compared with its ground-truth label. Performance is evaluated using accuracy, precision, recall, and F1-score. Accuracy measures the overall proportion of correctly classified samples, while precision and recall quantify class-specific predictive performance. The F1-score provides their harmonic mean and is useful when a balanced assessment of precision and recall is desired.

### 3.5 Overall Proposed Framework

The complete proposed framework therefore consists of two parallel feature-extraction paths followed by feature-level concatenation and classification. For an input texture image, *I*, the DWT branch produces a 192-dimensional vector and the modified AlexNet branch produces a 256-dimensional vector. These vectors are concatenated to form a 448-dimensional representation, which is subsequently classified by the DNN–SoftMax classifier. In compact form, the overall mapping can be expressed as

$$I = \{\mathbf{f}_{DWT}, \mathbf{f}_{Alex}\} \rightarrow \mathbf{f}_F \in \mathbb{R}^{448} \rightarrow DNN \rightarrow SoftMax \rightarrow \hat{y} \tag{11}$$

The framework enables a direct comparison among DWT-DNN, AlexNet-DNN, and fused representations. The performance difference among these three configurations provides an empirical measure of the complementary information contributed by the two feature-extraction techniques. Improved performance of DWT-AlexNet-DNN model would indicate that the frequency-domain statistical descriptors and learned deep features provide complementary information for texture discrimination.

**Table 2.** Layers of DNN

| Layer | Type | Number of Neurons | Function |
|---|---|---|---|
| Input | Feature Input | 448 | Receives fused DWT + AlexNet features |
| FC1 | Fully Connected | 256 | Learns nonlinear combinations of fused features |
| ReLU1 | ReLU | 256 | Introduces nonlinearity |
| Dropout1 | Dropout | — | Reduces overfitting |
| FC2 | Fully Connected | 128 | Learns higher-level feature representations |
| ReLU2 | ReLU | 128 | Nonlinear activation |
| Dropout2 | Dropout | — | Regularization |
| FC3 | Fully Connected | 64 | Further feature abstraction |
| ReLU3 | ReLU | 64 | Nonlinear activation |
| FC4 | Fully Connected | 10 | Produces one score for each class |
| SoftMax | SoftMax | 10 | Converts scores into class probabilities |
| Output | Classification | 10 classes | Predicts the texture class |

# 4. Implementation and Results

To evaluate the effectiveness of the proposed feature-fusion approach, three classification models—DWT-DNN, AlexNet-DNN, and DWT-AlexNet-DNN—described in Section 3 were implemented in MATLAB. The DWT-DNN model classifies texture images using only the 192 DWT-based features extracted from each image. The AlexNet-DNN model uses a 256-dimensional feature vector extracted from a pretrained AlexNet network. The DWT-AlexNet-DNN model employs feature-level fusion by concatenating the 192 DWT features with the 256 AlexNet features, resulting in a 448-dimensional fused feature vector. The performance of the three models was evaluated on three widely used benchmark texture datasets: Brodatz, KTH-TIPS, and the Flickr Material Database (FMD). To ensure a consistent and fair comparison, each dataset was partitioned into 70% training, 15% validation, and 15% testing subsets. The same partitioning strategy was applied to all three models for each dataset. The characteristics of the datasets, experimental procedures, and classification results are presented in the following subsections.

## 4.1 Brodatz Dataset

The Brodatz texture dataset is a widely used benchmark for evaluating texture analysis and classification algorithms. Originally introduced by Brodatz, the dataset contains 54 grayscale texture images representing a diverse range of natural and man-made surface patterns, including fabrics, wood, sand, leather, stone, and other materials [25]. Each texture exhibits distinctive spatial and structural characteristics, making the dataset suitable for evaluating the ability of image-processing and machine-learning methods to discriminate among different texture classes. The Brodatz dataset has been extensively used in studies involving handcrafted texture descriptors, wavelet-based methods, local binary patterns, filter-bank approaches, and deep-learning techniques. In this work, DWT-based features were extracted from the 54 Brodatz texture images. The extracted features were subsequently clustered into three groups—granular, coarse, and patterned—using the K-means clustering algorithm. These three clusters are treated as the three classification classes. The resulting clusters were used to organize the dataset for the experimental evaluation of the proposed framework. The hybrid approach combines multiresolution DWT-based texture features with deep features extracted using the modified AlexNet architecture described in Section 3. Representative sample images from the three clusters are shown in Fig. 4. The Brodatz dataset was evaluated using all three classification models: the DWT-DNN model, the AlexNet-DNN model, and the proposed fused-feature model. The training progress and confusion matrix obtained using the DWT-AlexNet-DNN model are presented in Figs. 5 and 6, respectively. A comparative evaluation based on overall accuracy, precision, recall, and F1-score is presented in Table 3, and the corresponding results are illustrated in the bar chart in Fig. 7. As observed from Fig. 7, the DWT-DNN and DWT-AlexNet-DNN models achieve comparable performance across the evaluated metrics for the Brodatz dataset.

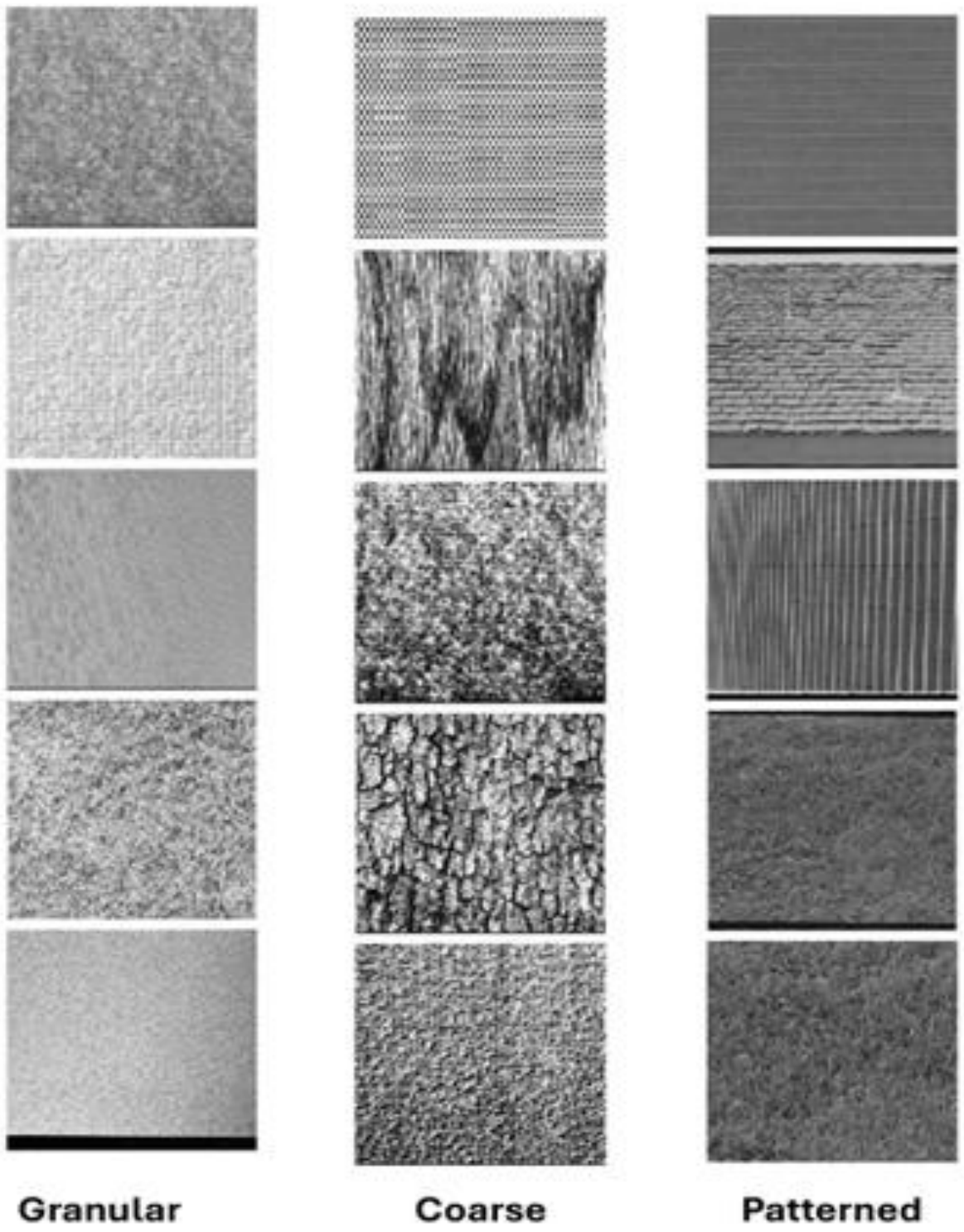

**Fig. 4**. Sample images from the Brodatz Dataset

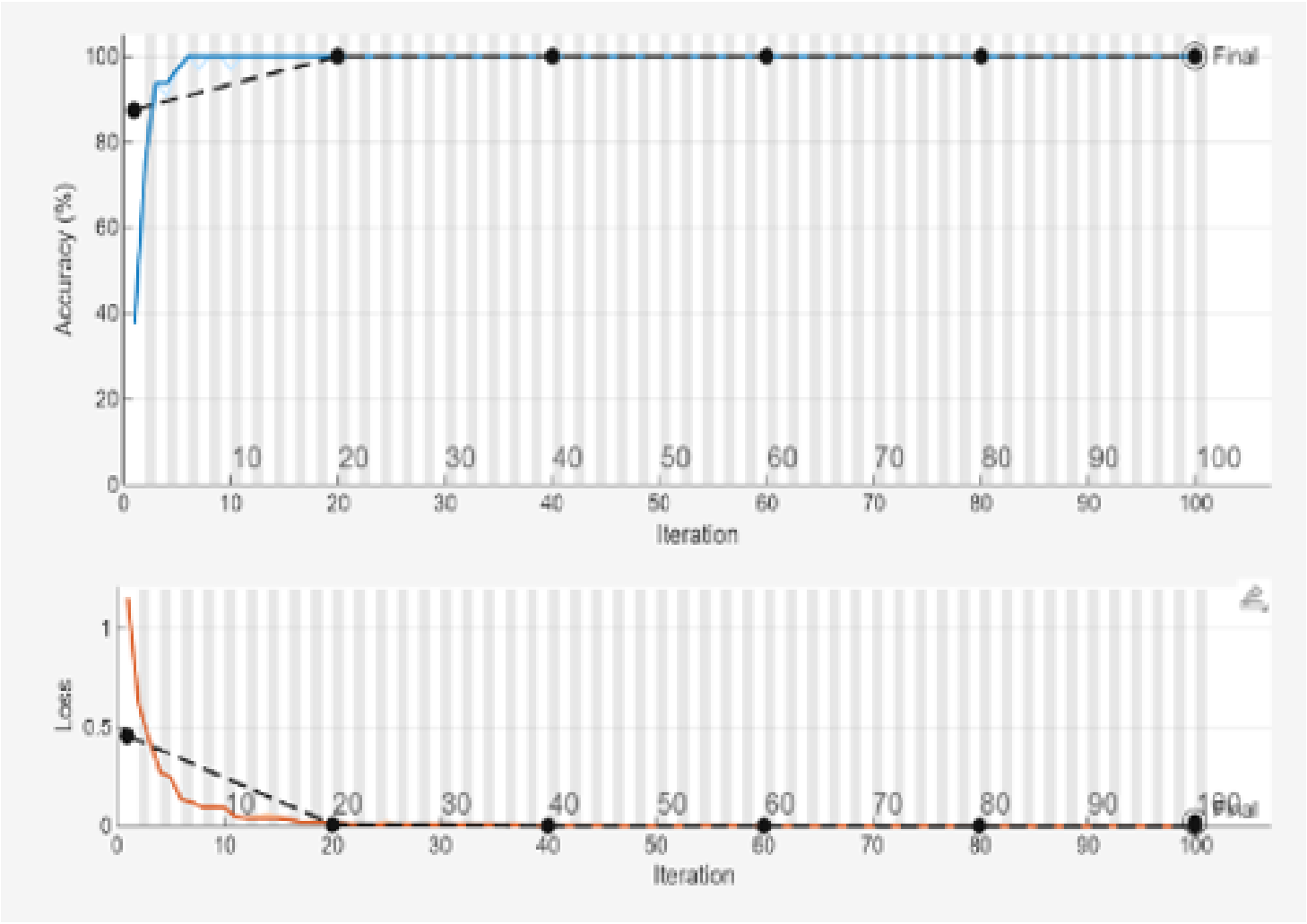

**Fig. 5**. Training progress of the DWT-AlexNet-DNN model Brodatz dataset

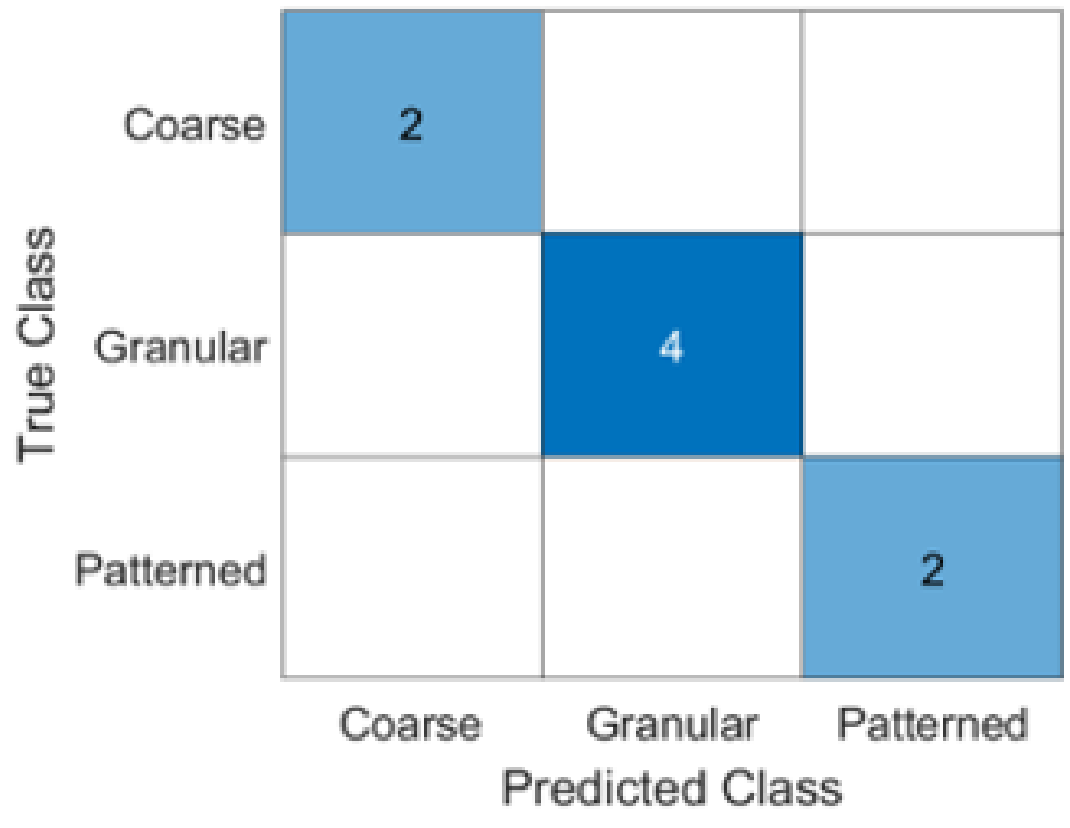

**Fig. 6**. Confusion Matrix of the DWT-AlexNet-DNN model - Brodatz dataset

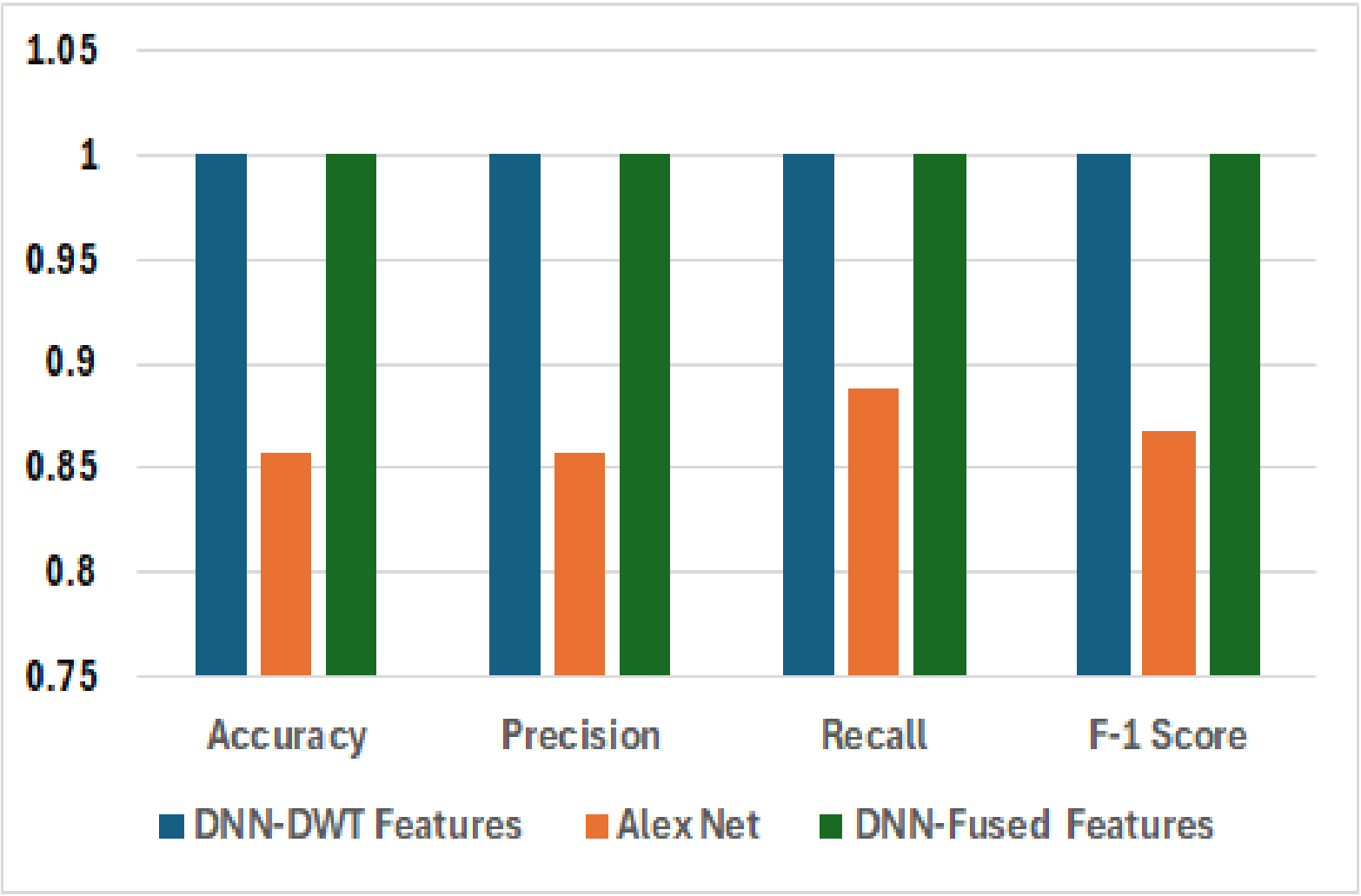

**Fig. 7.** Results for Brodatz dataset

**Table 3**. Classification Results for Brodatz Dataset

| | Accuracy | Precision | Recall | F1-score |
|---|---|---|---|---|
| DWT-DNN | 1.0 | 1.0 | 1.0 | 1.0 |
| AlexNet-DNN | 0.8571 | 0.857 | 0.889 | 0.867 |
| DWT-AlexNet-DNN | 1.0 | 1.0 | 1.0 | 1.0 |

## 4.2. KTH-TIPS Dataset

The KTH-TIPS (Textures under varying Illumination, Pose, and Scale) dataset is a widely used benchmark for evaluating texture classification and recognition methods under variations in imaging conditions. The dataset contains images of real-world materials acquired under different illumination conditions, scales, and viewing angles [26]. These variations make KTH-TIPS particularly suitable for assessing the robustness and generalization capability of texture descriptors and classification models. The KTH-TIPS dataset contains 810 images distributed among 10 texture classes, with 81 images per class. Consequently, images belonging to the same material class may exhibit substantial visual variation, whereas images from different classes may contain similar local patterns. These characteristics present a challenging classification problem and provide an appropriate testbed for evaluating both handcrafted and deep-learning-based texture descriptors. KTH-TIPS has been extensively used to evaluate handcrafted texture descriptors, filter-bank and wavelet-based methods, convolutional neural networks, and other deep-learning approaches. In this work, the KTH-TIPS dataset was used to evaluate the proposed hybrid feature-fusion framework, which combines multiresolution DWT-based features with deep features extracted using the modified AlexNet architecture. The performance of the proposed fused-feature model was compared with that of the DWT-DNN and AlexNet-DNN models using overall accuracy, precision, recall, and F1-score. Representative sample images from the ten classes are shown in Fig. 8. The training progress and confusion matrix for the DWT-AlexNet-DNN model are presented in Figs. 9 and 10, respectively. A comparative evaluation based on overall accuracy, precision, recall, and F1-score is presented in Table 4, and the corresponding bar chart representation is shown in Fig. 11. As observed from Fig. 11, the fused-feature model achieves better overall performance than the models based solely on DWT or AlexNet features across the evaluated metrics. This improvement indicates that the complementary information provided by the handcrafted DWT features and the deep AlexNet features contribute to more effective texture discrimination.

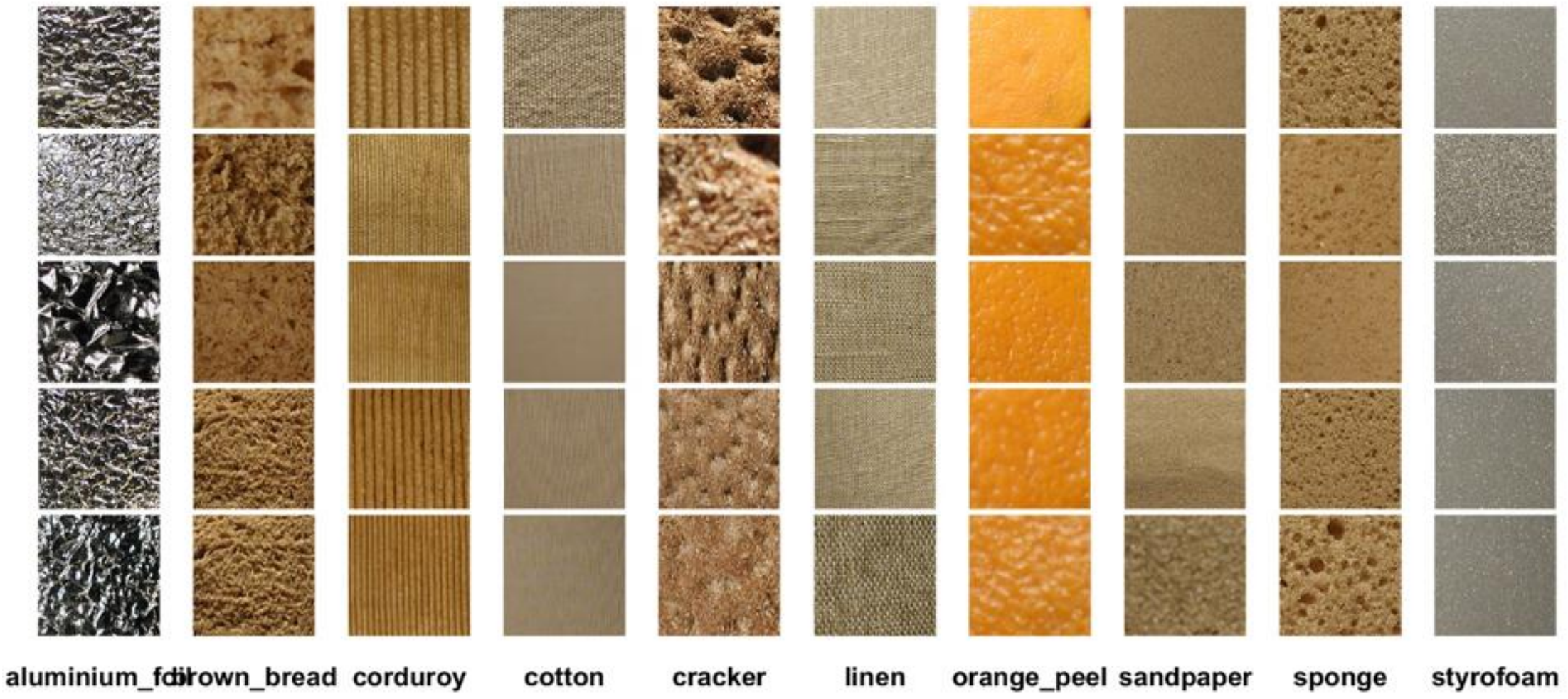

**Fig. 8.** Sample Images KTH-TIPS dataset

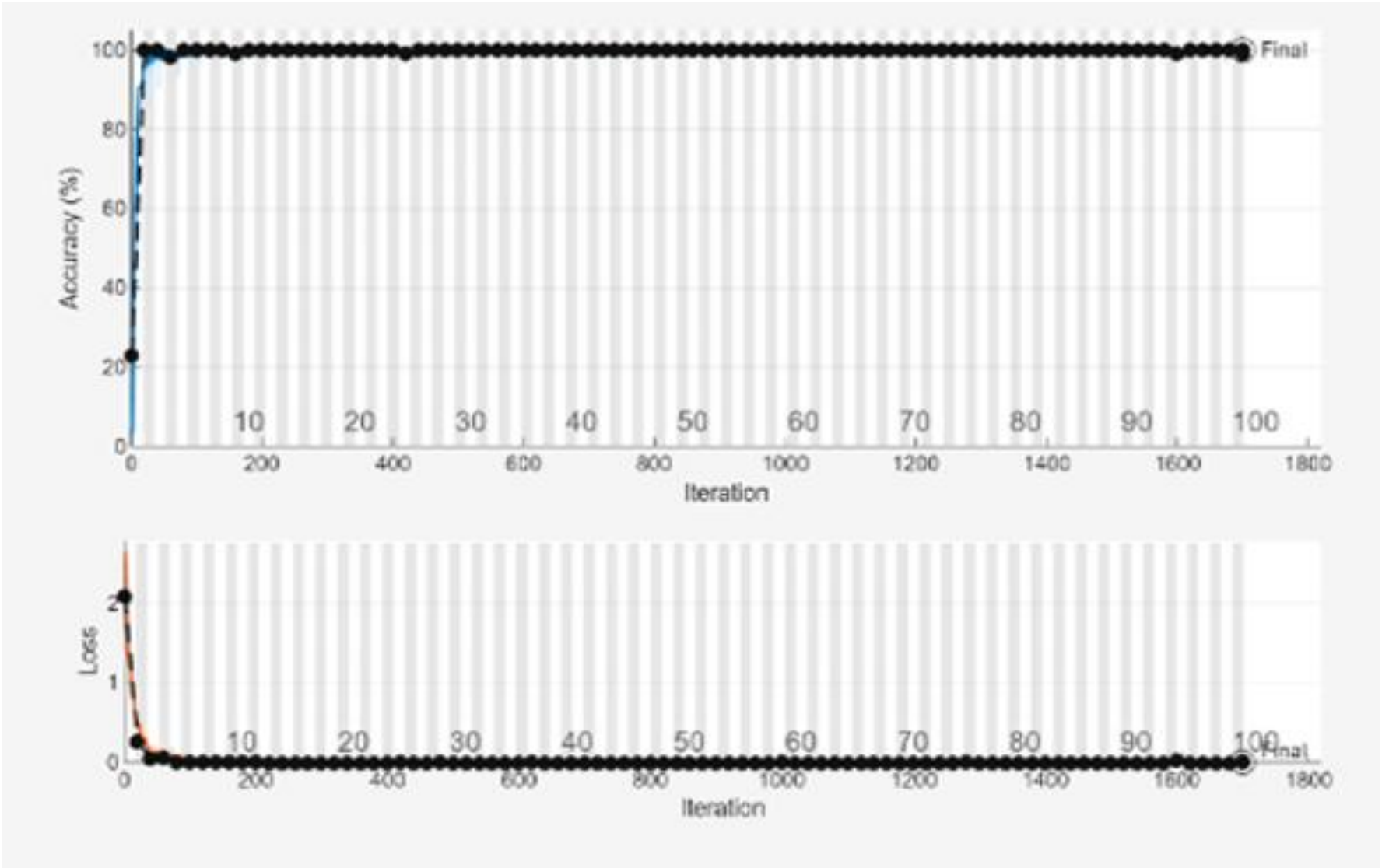

**Fig. 9**. Training progress of the DWT-AlexNet-DNN model -KTH-TIPS dataset

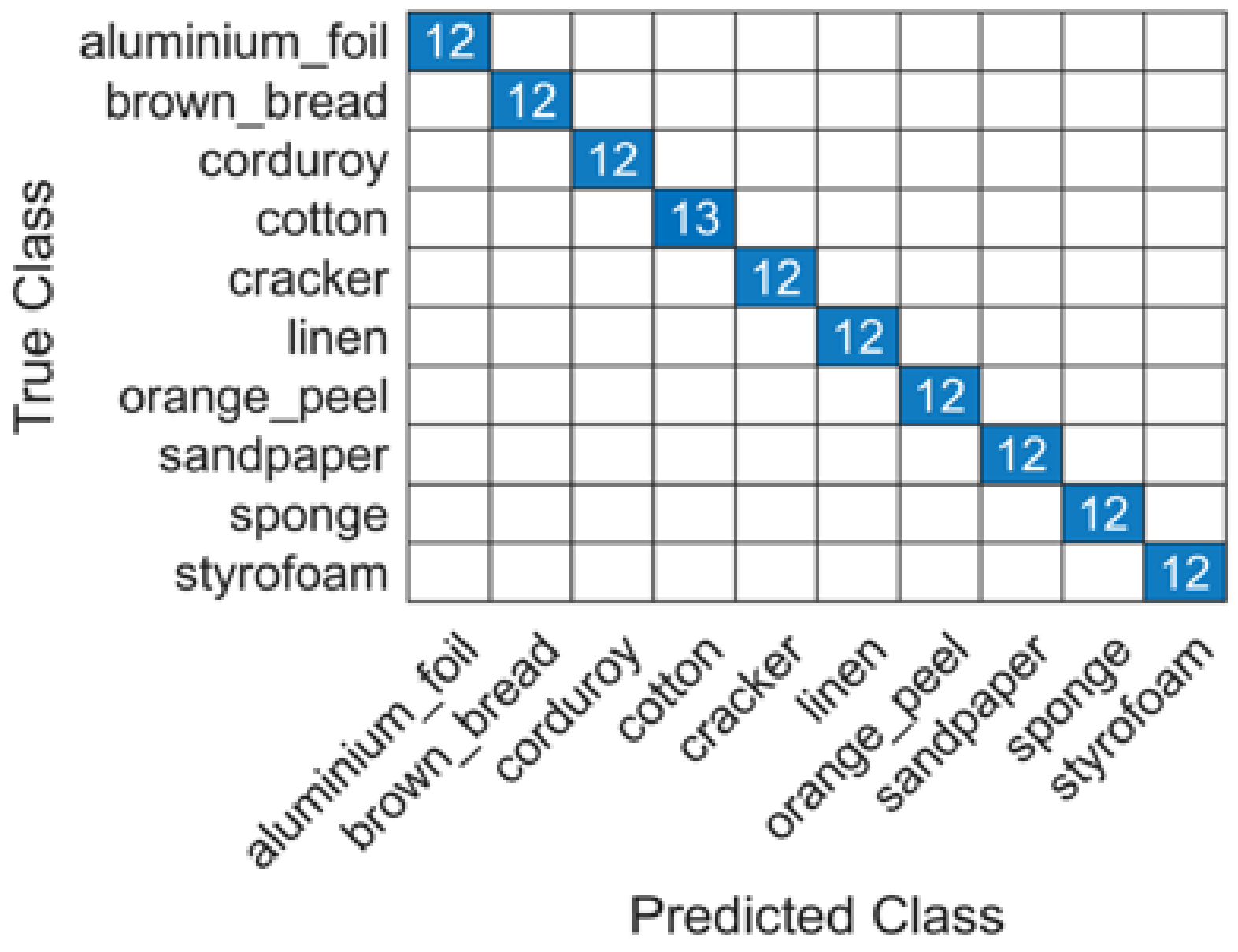

**Fig. 10.** Confusion Matrix DWT-AlexNet-DNN model on the KTH-TIPS dataset

**Table 4.** Classification Results for KTH-TIPS Dataset

| | Accuracy | Precision | Recall | F1-score |
|---|---|---|---|---|
| DWT-DNN | 0.9587 | 0.959 | 0.958 | 0.958 |
| AlexNet-DNN | 0.9917 | 0.992 | 0.992 | 0.992 |
| DWT-AlexNet-DNN | 1.0 | 1.0 | 1.0 | 1.0 |

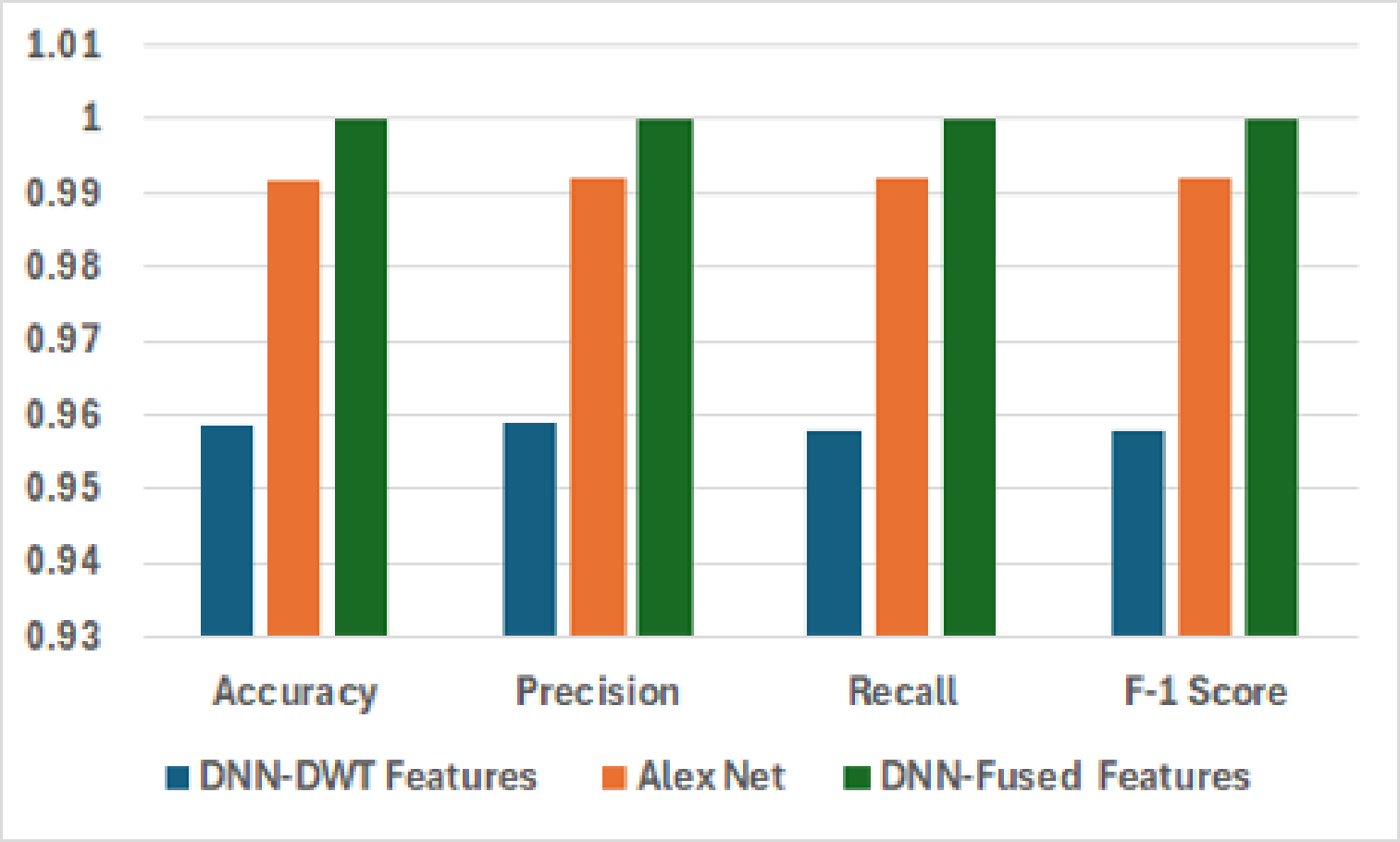

**Fig. 11.** Results for KTH-TIPS dataset

### 4.3 Flickr Material Dataset (FMD)

The Flickr Material Database (FMD) is a widely used benchmark dataset for material and texture recognition in computer vision. It consists of real-world images collected from Flickr and covers a diverse range of material categories, including fabric, glass, leather, metal, paper, plastic, stone, and

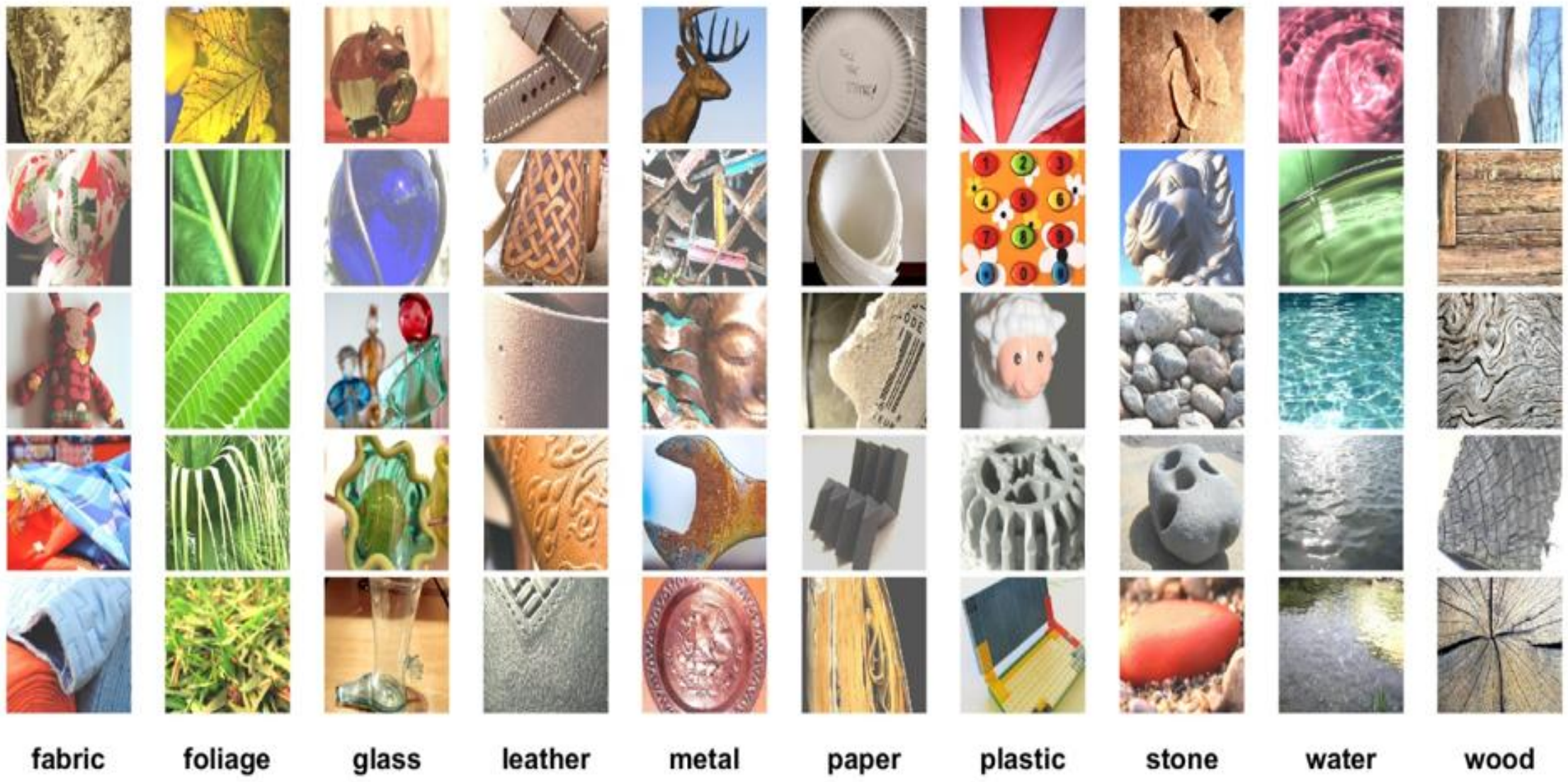

**Fig. 12.** Sample Images from the FMD dataset

wood. Unlike controlled laboratory texture datasets [27], FMD contains images acquired under varying illumination, viewpoint, scale, and background conditions. Consequently, the dataset exhibits substantial intra-class variability and inter-class similarity, making it particularly suitable for evaluating the robustness and generalization capability of texture and material classification algorithms. FMD has been widely used to evaluate handcrafted texture descriptors, filter-bank and wavelet-based methods, and deep-learning approaches for material recognition. The FMD dataset consists of 1,000 images distributed across 10 material categories, with 100 images in each category. In this work, the FMD dataset was used to evaluate the proposed hybrid feature-fusion framework, which combines multiresolution DWT-based texture features with deep features extracted using the modified AlexNet architecture. The classification performance of the proposed fused-feature model was compared with the corresponding DWT-DNN and AlexNet-DNN models using standard evaluation metrics, including accuracy, precision, recall, and F1-score. Representative sample images from the ten classes are shown in Fig. 12. The training progress and confusion matrix obtained using the DWT-AlexNet-DNN model are presented in Figs. 13 and 14, respectively. A comparative evaluation based on overall accuracy, precision, recall, and F1-score is presented in Table 5, while the corresponding bar chart representation is shown in Fig. 15. As observed from Fig. 15, the fused-feature model achieves better overall performance than the models using only DWT or AlexNet features across the evaluated metrics. These results demonstrate the effectiveness of feature-level fusion in combining

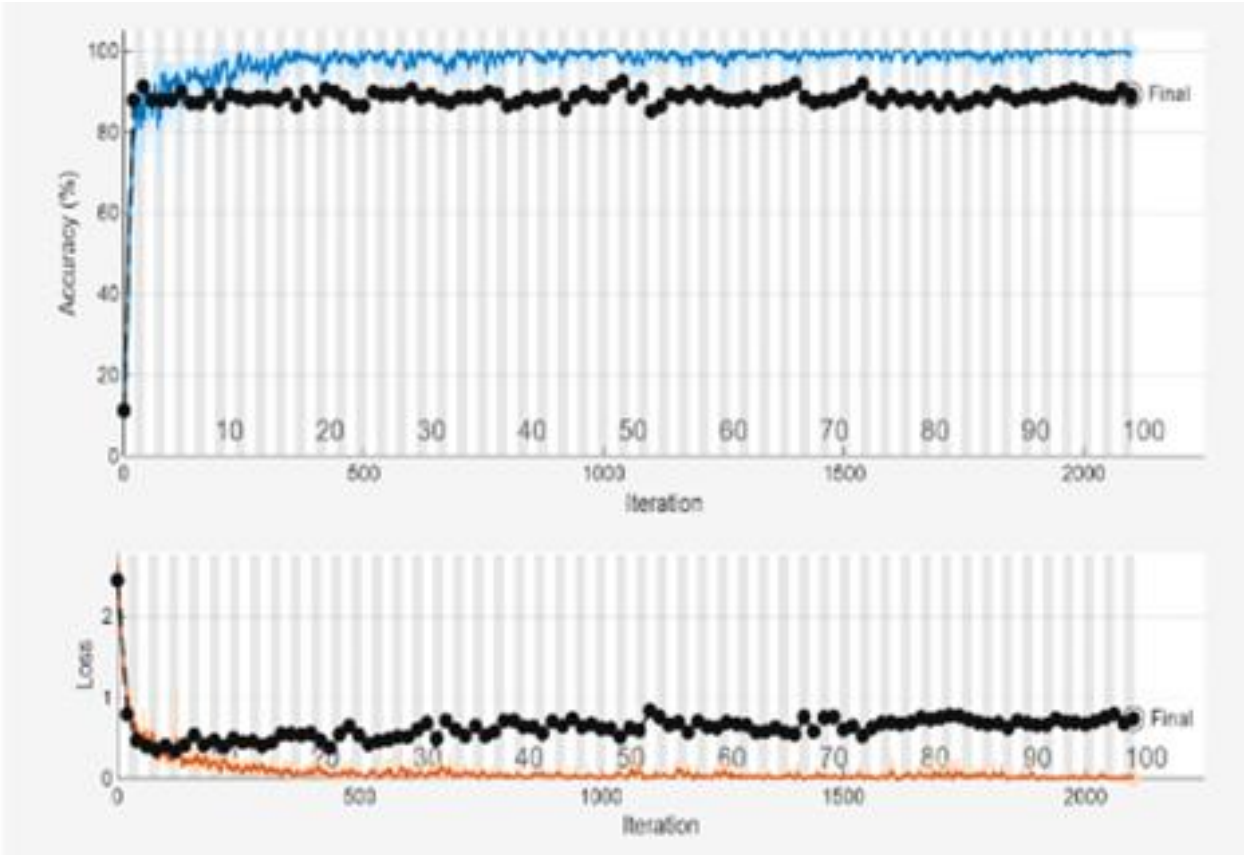

**Fig. 13.** Training progress of the DWT-AlexNet-DNN model on the FMD dataset

complementary multiresolution texture information from DWT with high-level discriminative representations learned by AlexNet

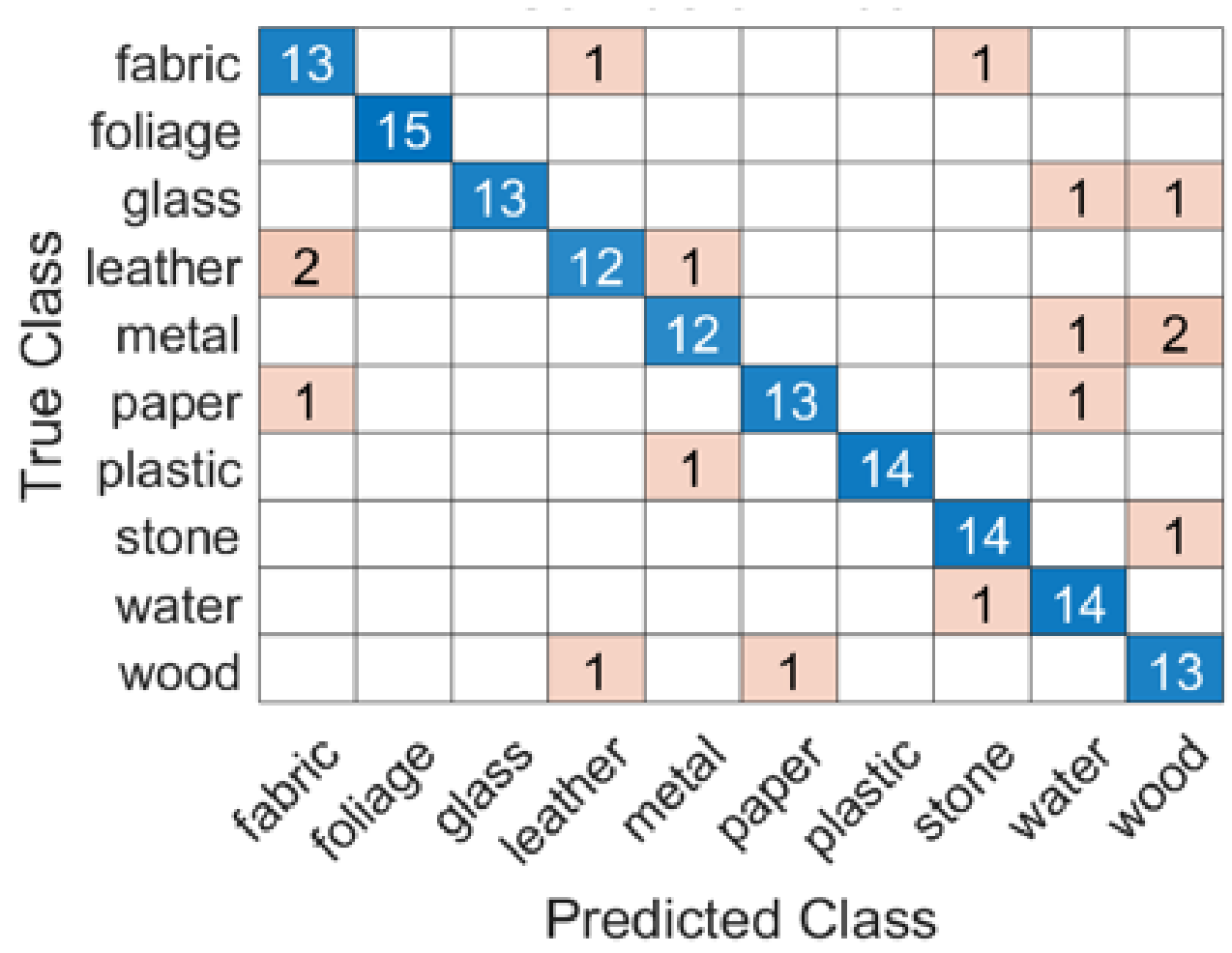

**Fig. 14.** Confusion Matrix of the DWT-AlexNet-DNN model FMD dataset

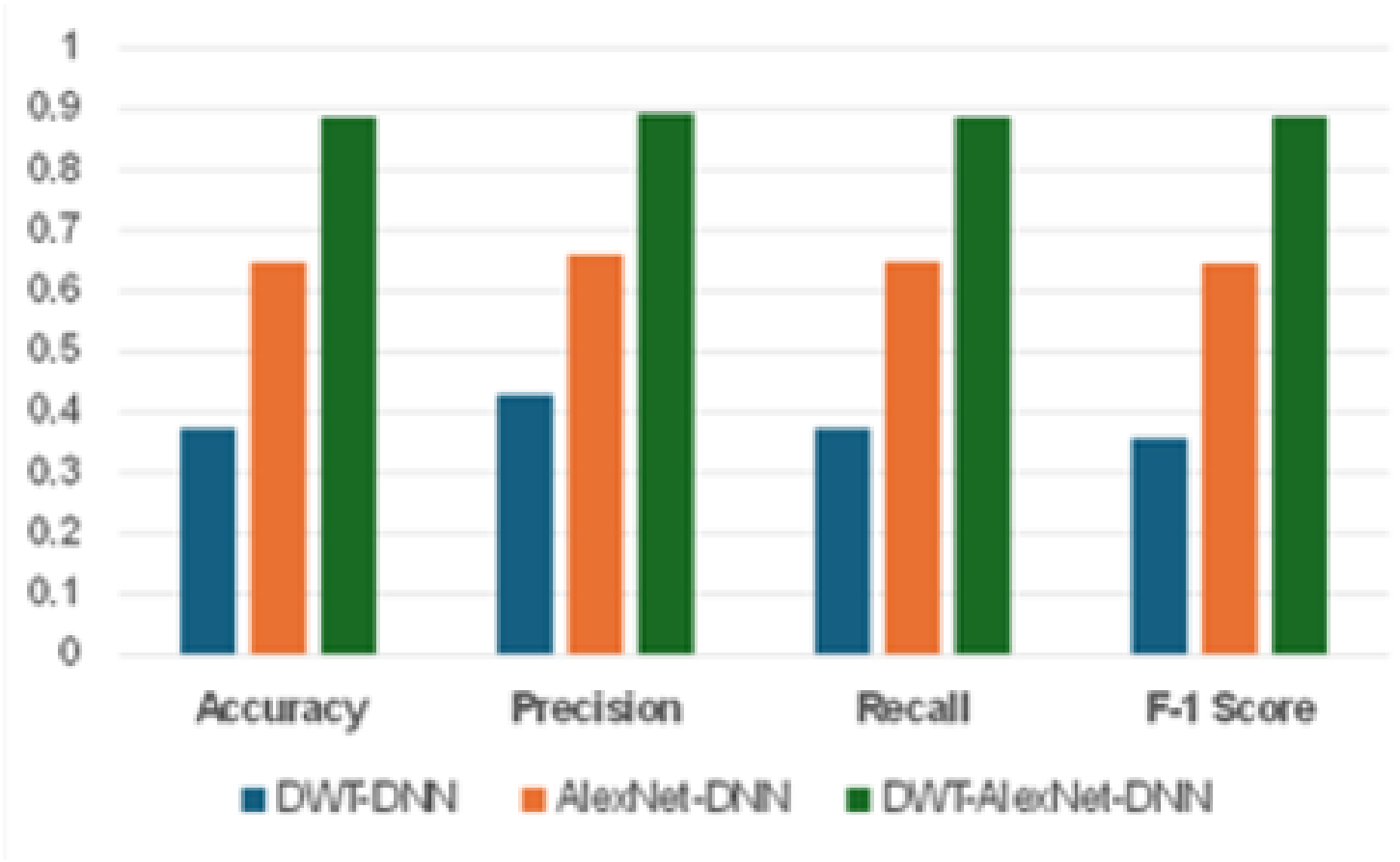

**Fig. 15.** Results for the FMD dataset

**Table 5.** Classification Results for FMD Dataset

| | Accuracy | Precision | Recall | F1-score |
|---|---|---|---|---|
| DWT-DNN | 0.3733 | 0.428 | 0.373 | 0.357 |
| Alex Net-DNN | 0.6467 | 0.659 | 0.647 | 0.645 |
| DWT-AlexNet-DNN | 0.8867 | 0.8919 | 0.8867 | 0.8875 |

## 5. Conclusion and Future Work

In this work, we proposed a hybrid feature-fusion model, referred to as the DWT-AlexNet-DNN model, for texture image classification. The proposed approach combines handcrafted frequency-domain descriptors with deep features learned by a convolutional neural network. The framework consists of two parallel feature-extraction paths. The first path extracts texture descriptors using the Discrete Wavelet Transform (DWT), while the second path extracts a 256-dimensional deep feature vector from a pretrained AlexNet model. The two feature vectors are subsequently concatenated to form a unified representation of each texture image. The resulting 448-dimensional fused feature vector is classified using a deep neural network (DNN) consisting of fully connected layers followed by a SoftMax classification layer. To evaluate the effectiveness of the proposed feature-fusion strategy, two additional models, DWT-DNN and AlexNet-DNN, were implemented for comparison. This experimental design provides a direct comparison of the three feature representations and enables the contribution of feature fusion to be quantitatively assessed. The proposed framework was evaluated using three benchmark texture datasets: Brodatz, KTH-TIPS, and the Flickr Material Database (FMD). The experimental results presented in Section 4 demonstrate that the effectiveness of feature fusion depends on the characteristics of each dataset. KTH-TIPS and FMD contain images of real-world materials acquired under variations in material appearance, color, texture, illumination, viewing angle, and imaging distance, making them useful benchmarks for evaluating the robustness of texture classification methods. KTH-TIPS contains 10 material categories with 81 images per category, whereas FMD contains 10 material categories with 100 images per category. All experiments were conducted in MATLAB on a system equipped with an AMD Ryzen 9 7950X processor and 128 GB of RAM.

For the Brodatz dataset, the results shown in Fig. 7 indicate that the DWT-DNN and DWT-AlexNet-DNN models achieve comparable performance across the evaluated metrics. This result suggests that the handcrafted DWT descriptors already capture a substantial portion of the discriminative information required for texture classification in this dataset. Consequently, incorporating the AlexNet features does not provide a significant additional improvement in classification performance. This observation indicates that, for homogeneous texture datasets, a frequency-domain representation may already provide sufficient information for effective classification. For KTH-TIPS, the AlexNet-DNN model achieves 99.17% accuracy, whereas the proposed DWT-AlexNet-DNN model achieves 100% accuracy. For Brodatz, both the DWT-DNN and DWT-AlexNet-DNN models achieve 100% accuracy. This represents an absolute improvement of 0.83 percentage points and a relative improvement of approximately 0.84%. Although the numerical improvement is small because the AlexNet-based representation already provides remarkably high classification accuracy, the fused representation achieves perfect classification on the evaluated test set. This result suggests that the DWT features provide complementary information even when the deep features are already highly discriminative. A substantially larger improvement is observed for the FMD dataset. As shown in Fig. 15, the classification accuracy increases from 64.67% using the AlexNet-DNN model to 88.67% using the proposed DWT-AlexNet-DNN model. This corresponds to an absolute improvement of 24.00 percentage points and a relative improvement of approximately 37.11%. The substantial improvement demonstrates the effectiveness of combining complementary handcrafted and deep representations, particularly for datasets containing greater variations in material appearance and imaging conditions. The results further indicate that frequency-domain texture descriptors can provide information that is not fully captured by the deep features extracted from pretrained AlexNet.

Overall, the experimental results demonstrate that handcrafted DWT descriptors and AlexNet deep features capture complementary characteristics of texture images. The DWT features provide frequency-domain information associated with local texture patterns at multiple resolutions, whereas the deep features extracted by AlexNet provide higher-level discriminative representations learned from image data. The improved performance obtained through feature fusion, particularly for the FMD dataset, indicates that these complementary representations can be effectively integrated to construct a more informative feature space for texture classification. The proposed framework therefore provides a flexible approach for combining handcrafted and learned features without requiring substantial modifications to the underlying feature-extraction models. The results also demonstrate that feature fusion does not necessarily provide the same level of improvement across all datasets. Its effectiveness depends on the diversity and complexity of the image content and on the extent to which the individual feature representations capture complementary information. For relatively homogeneous datasets, a single feature representation may already provide sufficient discriminative information. In contrast, for datasets characterized by substantial variations in material appearance and imaging conditions, combining frequency-domain and deep representations can provide a more comprehensive description of the underlying texture characteristics.

Future work will investigate the integration of features extracted from other deep convolutional neural network architectures, such as VGG-16, ResNet, and GoogLeNet, with handcrafted descriptors, including DWT and gray-level co-occurrence matrix (GLCM) features. The proposed approach will also be evaluated on larger and more diverse datasets containing a greater number of categories and substantially more training and testing images. Further investigation will focus on feature-selection and dimensionality-reduction techniques to determine whether the fused representation can be made more compact while maintaining or improving classification performance. Such techniques may reduce the computational and storage requirements associated with the higher-dimensional fused feature vectors.

In addition, the proposed framework could be extended to practical applications involving texture and material recognition, including industrial inspection, medical image analysis, remote sensing, and object and material recognition. Future investigations will also examine the computational complexity and scalability of the framework and evaluate its generalization to previously unseen datasets and imaging conditions. These studies will provide further insight into the scalability, computational efficiency, and generalization capability of the proposed feature-fusion approach and help determine its suitability for real-world texture classification applications.

**Data Availability:** All three datasets used in this research work are available in public domain at the following websites.

**Brodatz Texture Dataset**. Kaggle.
https://www.kaggle.com/datasets/whonosefirst/brodatz-texture-dataset
**KTH-TIPS database**. https://www.nada.kth.se/cvap/databases/kth-tips.
**Flickr Material Database** (FMD) (n.d.) https://people.csail.mit.edu/lavanya/fmd